\documentclass[10pt,twocolumn,a4paper]{article}

\usepackage[numbers]{natbib}
\usepackage{threeparttable}
\usepackage{tikz}
\usepackage{pgfplots}
\usepackage{booktabs}
\usepackage{enumitem}
\usepackage[margin=1.5cm]{geometry}
\setlist[itemize]{nosep, leftmargin=*}
\setlist[enumerate]{nosep, leftmargin=*}
\usepackage{authblk}
\usepackage[colorlinks=true,linkcolor=blue,citecolor=blue,urlcolor=blue]{hyperref}
\usepackage{microtype}
\usepackage{ragged2e}
\usepackage{multirow}
\usepackage{pgfplots}
\usepackage{pgfplotstable} % Per gestire meglio i file CSV
\usetikzlibrary{matrix}
\usepgfplotslibrary{groupplots}
\usetikzlibrary{shadows}
\pgfplotsset{compat=newest}

\usepackage{amsmath}
\usepackage{amssymb}

\usepgfplotslibrary{external}
\pgfplotsset{
    plotstyle/.style={
        width=\columnwidth,       % Larghezza pari a una colonna
        height=6cm,               % Altezza proporzionata
        grid=major,               % Griglia principale
        grid style={dashed, gray!30},
        tick label style={font=\scriptsize},
        label style={font=\scriptsize},
        title style={font=\scriptsize},
    }
}

\pgfplotscreateplotcyclelist{graph_precision_recall}{
    {blue, no marks, solid, thick},            % SBMEX
    {blue, no marks, dotted, thick},           % SBMEX-R    
    {red!70!black, only marks, mark=triangle*, mark size=1.5pt, thick}, % CNBaseline: punto
    {red!70!black, no marks, solid, thick},           % MINDTCT (Raw)
    {red!70!black, no marks, dashed, thick},          % MINDTCT (Enh)
    {green!70!black, no marks, solid, thick},         % FingerJet (Raw)
    {green!70!black, no marks, dashed, thick},        % FingerJet (Enh)        
    {black!60, no marks, solid, thick},               % VeriFinger (Raw)
    {black!60, no marks, dashed, thick},              % VeriFinger: (Enh)
    {black, no marks, solid, thick},               % LEADER (Raw)
    {black, no marks, dashed, thick},              % LEADER: (Enh)
}

\newcommand{\gridcell}[4]{%
  \ifnum#3=0
    \fill[black] (#1,#2) rectangle ++(1,1);
    \node[white, font=\tiny] at (#1+0.5, #2+0.5) {#4};
  \else
    \node[black!60, font=\tiny] at (#1+0.5, #2+0.5) {#4};
  \fi
}

\newcommand{\gridthree}[9]{%
  \begin{tikzpicture}[scale=0.22, baseline={(0,0.2)}]
    \gridcell{0}{0}{#7}{6}
    \gridcell{1}{0}{#8}{5}
    \gridcell{2}{0}{#9}{4}
    \gridcell{2}{1}{#6}{3}
    \gridcell{1}{1}{#5}{}
    \gridcell{2}{2}{#3}{2}
    \gridcell{1}{2}{#2}{1}
    \gridcell{0}{2}{#1}{0}
    \gridcell{0}{1}{#4}{7}
    \draw[black!50, thick, step=1cm] (0,0) grid (3,3);
  \end{tikzpicture}%
}

\begin{document}

\title{Revisiting Handcrafted Minutiae Detection: A Simple and Effective\\ Open Source Baseline for Modern Fingerprint Workflows}
\author{Raffaele Cappelli}
\affil{Department of Computer Science and Engineering, University of Bologna, Cesena, Italy\\
\href{mailto:raffaele.cappelli@unibo.it}{\texttt{raffaele.cappelli@unibo.it}}}
\date{} % Rimuove la data automatica

\twocolumn[
  \begin{@twocolumnfalse}
    \maketitle
        \begin{abstract}
        Handcrafted minutiae detection algorithms remain fundamental to biometric science and forensic practice due to their full auditability, adherence to international standards, and operational independence from training datasets or GPU hardware. However, current open-source traditional baselines are severely outdated, relying almost exclusively on legacy C/C++ codebases that lack seamless integration with modern scientific software ecosystems. To bridge this gap, the present work introduces SBMEX (Skeleton-Based Minutiae EXtraction), a fast and deterministic minutiae detection method integrated into the open source \texttt{pyfing} package. SBMEX achieves high computational throughput by employing a dual Look-Up Table architecture that replaces runtime neighborhood scanning during Crossing Number computation and skeleton tracking. Additionally, it incorporates a continuous quality scoring framework driven by tracking path length, dual ridge-valley skeleton fusion, and spatial density decay. Rigorous evaluation on NIST SD302 datasets demonstrates that SBMEX delivers feature extraction accuracy comparable to or outperforming traditional open-source baselines without fine-tuning, while achieving a drastic reduction in minutiae detection latency relative to classical Crossing Number Python implementations.
        \end{abstract}
    \vspace{0.5em}
    \vspace{1.5em}
  \end{@twocolumnfalse}
]

\maketitle

\section{Introduction}
\label{sec:introduction}
\begin{figure}
	\centering
	\includegraphics[width=\columnwidth]{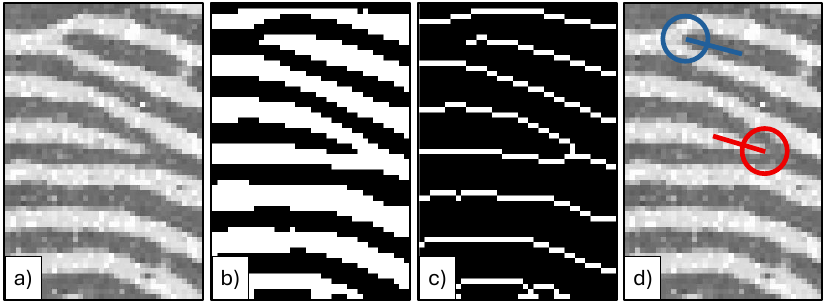}
	\caption{Classical minutiae extraction pipeline. (a) Portion of a fingerprint image displaying dark gray ridges separated by lighter valleys. (b) Enhanced and binarized image, where ridges are mapped to white pixels and valleys to black pixels. (c) Topological skeletonization of the ridges (1-pixel wide white paths). (d) Extracted minutiae features, highlighting a ridge ending near the top (blue circle) and a bifurcation located lower towards the center (red circle). The line segment originating from the center of each circle indicates the direction of the corresponding minutia, defined by an angle in \([0, 2\pi)\).}
	\label{fig:classic_pipeline}
\end{figure}

Fingerprint recognition remains one of the most reliable and widely deployed biometric technologies for personal identification in civilian, law enforcement, and forensic applications \cite{Maltoni2022}. Automated Fingerprint Identification Systems (AFIS) rely heavily on minutiae—local ridge discontinuities—as the primary distinctive features for representation and matching. Among various minutia types, ridge endings (points where a ridge abruptly terminates) and bifurcations (points where a ridge forks into two branches) constitute the vast majority of extracted features (see Fig.~\ref{fig:classic_pipeline}.d). Despite decades of advancement, accurate minutiae detection remains a challenging task, particularly when processing low-quality impressions such as latent fingermarks characterized by severe noise, incomplete ridge structures, and background interference \cite{Maltoni2022}.

In recent years, state-of-the-art performance in minutiae detection has been increasingly driven by deep neural networks \cite{Cappelli2026a}. Nevertheless, the traditional handcrafted minutiae detection pipeline, historically developed over more than half a century, retains paramount importance in biometric science and forensic practice. A typical classical pipeline (Fig.~\ref{fig:classic_pipeline}) operates sequentially through image enhancement, binarization, skeletonization, and minutiae detection via connectivity analysis. This deterministic paradigm offers some critical advantages:
\begin{itemize}
    \item \textit{Explainability and legal admissibility}: Handcrafted feature extraction directly mirrors the explicit, rule-based procedures employed by human latent examiners, providing full algorithmic transparency and auditability required in judicial settings where black-box deep learning models may face strict legal evidentiary hurdles.
    \item \textit{Standardization compliance}: International biometric standards, such as ISO/IEC 19794-2 \cite{iso19794-2}, define minutiae attributes (e.g., location, direction, and type) based on explicit geometric rules along skeletonized ridge lines, ensuring seamless interoperability across heterogeneous system architectures.
    \item \textit{Computational efficiency and zero-shot deployment}: Handcrafted algorithms execute deterministically without requiring extensive training datasets, high-end GPU hardware, or hyperparameter fine-tuning across domain shifts, enabling immediate zero-shot deployment on edge devices or resource-constrained environments.
\end{itemize}

Despite the ongoing necessity for traditional minutiae detection, the current landscape of open-source baselines for classical pipelines is severely outdated. The research community relies almost exclusively on legacy software tools, primarily NIST MINDTCT \cite{Watson2007} and FingerJet \cite{DigitalPersona}. While foundational, these packages present major operational drawbacks: they are written in legacy C/C++ codebases, are rigid to modify, and lack native integration with modern scientific software ecosystems, which are now predominantly Python-based. Consequently, researchers aiming to prototype or benchmark traditional workflows are forced to resort to inefficient subprocess wrappers around compiled binaries or to implement ad-hoc scripts from scratch; a practical constraint illustrated by Oblak et al. \cite{Oblak2022}, who had to integrate MINDTCT and FingerJet via Python wrappers alongside a custom Crossing Number (CN) implementation. The persistence of this limitation is evidenced by numerous very recent works \cite{Koop2026, Gundgurti2026, Wang2026, Houari2026}, which continue to depend on legacy tools like MINDTCT as their primary handcrafted baseline due to the absence of a modern alternative.

To bridge this gap, the present work introduces SBMEX (Skeleton-Based Minutiae EXtraction), a fast, deterministic, and fully auditable minutiae detection algorithm designed to provide a modern reference baseline for traditional fingerprint pipelines. The main contributions of this work are summarized as follows:
\begin{itemize}
    \item A novel \textit{dual LookUp-Table (LUT) architecture} that accelerates both CN computation and ridge tracking, drastically speeding up skeleton traversal without complex conditional branching.
    \item A \textit{continuous quality scoring} paradigm that replaces traditional binary hard-filtering. Initial minutia quality is scaled continuously with tracking path length and attenuated via exponential cluster decay and dual-skeleton (ridge-valley) agreement.
    \item A \textit{flexible detection framework} allowing lightweight ridge-only analysis or full dual-skeleton merging to adapt to processing constraints.
    \item An \textit{extensive zero-shot evaluation} on the challenging NIST SD302 benchmark, demonstrating that SBMEX consistently outperforms or matches traditional baselines without requiring domain-specific fine-tuning.
    \item A \textit{fully open-source Python implementation} integrated into the \texttt{pyfing} ecosystem, offering the research community an easily accessible, modular, and reproducible tool.
\end{itemize}

The remainder of this paper is organized as follows. Section~\ref{sec:related_work} reviews related work on traditional and learning-based minutiae detection algorithms. Section~\ref{sec:methodology} describes the mathematical formulation and structural mechanics of the proposed algorithm. Section~\ref{sec:experiments} presents the experimental setup, comparative evaluations on benchmark datasets, and performance analysis. Finally, Section~\ref{sec:conclusion} concludes the paper and outlines future directions.

\section{Related work}
\label{sec:related_work}
This section reviews the main paradigms in minutiae extraction, covering traditional frameworks, deep learning models, and the existing open-source traditional baselines against which the proposed SBMEX is evaluated.

\subsection{Traditional methods}
Traditional algorithms are broadly categorized into \textit{binarization-based} and \textit{direct gray-scale} approaches.

Binarization-based frameworks enhance the input image, segment ridge-valley patterns, and tipycally compute a one-pixel-wide skeleton. Minutiae are subsequently detected by applying local connectivity rules, such as the CN metric, defined as the total number of \(0 \to 1\) pixel transitions along the ordered \(3 \times 3\) neighborhood sequence~\cite{Maltoni2022}. Because binarization is sensitive to noise, these pipelines heavily rely on heuristic post-processing rules to eliminate spurious detections caused by scars, dry skin, or low contrast.

To avoid information loss and artifacts introduced by binarization and skeletonization, direct gray-scale methods extract minutiae directly from raw (or enhanced) intensity values. These algorithms typically rely on ridge-following strategies along local orientation vectors~\cite{Maio1997} or model minutiae as local discontinuities in the ridge flow~\cite{Fronthaler2008}.

\subsection{Deep learning-based methods}
Deep learning has significantly advanced minutiae extraction by replacing handcrafted heuristics with end-to-end learned feature representations.

\textit{Patch-based and Multi-stage Networks}. Early deep learning frameworks utilized sliding-window CNNs to classify local patches as minutia or non-minutia~\cite{Sankaran2014, Darlow2017}. To improve context awareness, multi-stage architectures introduced region proposal networks followed by refinement sub-networks to regress coordinates and directions~\cite{Tang2017a, Nguyen2018a, Zhou2020}. While accurate, these sequential pipelines suffer from high computational overhead.

\textit{Single-pass Fully Convolutional Pipelines}. State-of-the-art approaches deploy single-pass fully convolutional networks that process the entire image in a single forward pass~\cite{Tang2017, Nguyen2020, Feng2023, Cappelli2026a}. These architectures output dense feature maps encoding minutia location, direction, and type simultaneously, achieving high accuracy and computational efficiency even on low-quality fingerprints.

\subsection{Existing open-source traditional baselines}
\label{subsec:baselines}
To evaluate SBMEX within a rigorous and reproducible framework, its performance is compared against three representative open-source traditional algorithms, each embodying a distinct extraction paradigm.

Developed by NIST within the NBIS suite~\cite{Watson2007}, \textit{MINDTCT} relies on directional filtering for binarization, local connectivity analysis for candidate extraction, and heuristic rules to eliminate false minutiae.

\textit{CNBaseline} is the traditional algorithm in~\cite{Oblak2022}, which applies standard morphological skeletonization to a pre-binarized image and applies the CN to identifies candidate minutiae. A final distance-based filtering pass removes closely adjacent candidate pairs.

\textit{FingerJet}~\cite{DigitalPersona} is an open-source lightweight engine representative of direct gray-scale extraction methods, since code analysis reveals that it avoids explicit binarization and skeletonization.

\section{Proposed method: SBMEX}
\label{sec:methodology}
This section details SBMEX, a fast and deterministic minutiae detection method designed for standard fingerprint processing pipelines.

\subsection{High-level overview}
\label{subsec:overview}
Given an input grayscale image \(I(x,y) \in [0, 255]\) and a segmentation mask \(M(x,y) \in \{0, 1\}\), SBMEX executes in five sequential stages:
\begin{enumerate}
    \item \textit{Binarization and thinning:} The image \(I\) is binarized within \(M\) using a global threshold \(t_b\) to yield a binary ridge map \(B_R(x,y) = \mathbb{I}(I(x,y) \le t_b \land M(x,y) = 1)\), where \(\mathbb{I}(\cdot)\) is the indicator function. The parallel thinning algorithm of Guo and Hall~\cite{Guo1989} reduces \(B_R\) to a 1-pixel-wide skeleton \(S_R(x,y) \in \{0, 1\}\).
    \item \textit{Candidate detection and perimeter filtering:} Candidate minutiae are extracted on \(S_R\) using a branchless CN lookup, discarding spurious candidates located near the background boundary of \(M\) (Section~\ref{subsec:candidate_detection}).
    \item \textit{Skeleton tracking and direction estimation:} Validated candidates undergo LUT-accelerated ridge-following to compute their direction \(\theta \in [0, 2\pi)\) and initial quality score \(q_0 \in (0, 1]\) based on the tracked path length (Section~\ref{subsec:skeleton_tracking}).
    \item \textit{Dual ridge-valley fusion (optional):} To improve extraction in low-contrast areas, steps 1--3 can be mirrored on the inverted binary valley map \(B_V(x,y) = \mathbb{I}(I(x,y) > t_b \land M(x,y) = 1)\). Inverted valley minutiae are geometrically shifted and paired with ridge minutiae, penalizing candidates lacking dual-domain agreement (Section~\ref{subsec:dual_extraction}).
    \item \textit{Cluster-based quality decay:} Final quality scores are continuously attenuated via spatial density decay to down-weight spurious minutiae clusters (Section~\ref{subsec:cluster_suppression}).
\end{enumerate}

The global parameters governing SBMEX are summarized in Table~\ref{tab:sbmex_params} and detailed in the following subsections.

\begin{table}[t]
\centering
\caption{Summary of SBMEX parameters and default values at $500$\,DPI.}
\label{tab:sbmex_params}
\small
\setlength{\tabcolsep}{4pt}
\begin{tabular}{llcc}
\toprule
Symbol & Description & Default & Unit \\
\midrule
\(t_b\) & Global binarization threshold & $0$ & -- \\
\(d_{\text{bg}}\) & Minimum distance to mask border & $10$ & px \\
\(d_{\text{min}}\) & Minimum tracking distance & $5$ & px \\
\(d_{\text{max}}\) & Maximum tracking distance & $20$ & px \\
\(d_{\text{pair}}\) & Max pairing distance & $6$ & px \\
\(\theta_{\text{pair}}\) & Max pairing angle difference & \(\pi/6\) & rad \\
\(w_{\text{unpair}}\) & Penalty for unpaired minutiae & $0.5$ & -- \\
\(\delta_{\text{val}}\) & Spatial shift for valley minutiae & $3$ & px \\
\(\sigma_{\text{cl}}\) & Interaction radius for clusters & $11$ & px \\
\(\gamma_{\text{cl}}\) & Quality decay base factor & $0.5$ & -- \\
\bottomrule
\end{tabular}
\end{table}

\subsection{Candidate detection and perimeter filtering}
\label{subsec:candidate_detection}
To avoid runtime conditional loops, SBMEX encodes the 8-neighborhood of each skeleton pixel into an 8-bit integer state \(E(x,y) \in [0, 255]\) via 2D spatial cross-correlation:
\begin{equation}
    E(x,y) = \sum_{k=0}^{7} 2^k \cdot S_R(x + \Delta x_k, y + \Delta y_k)
\end{equation}
where \((\Delta x_k, \Delta y_k)\) denote the spatial offset vectors for the 8-neighborhood, ordered clockwise starting from \((-1, -1)\). Formulating this step as a standard 2D correlation allows the algorithm to leverage highly optimized routines from mainstream scientific computing libraries.

The CN value for each pixel \(CN(x,y)\) is subsequently retrieved in \(O(1)\) time using a pre-computed lookup table \(\text{LUT}_{CN}\):
\begin{equation}
    CN(x,y) = \begin{cases} \text{LUT}_{CN}[E(x,y)] & \text{if } S_R(x,y) = 1 \\ 0 & \text{otherwise} \end{cases}
\end{equation}
Pixels with \(CN(x,y) = 1\) are candidate ridge endings, while \(CN(x,y) = 3\) identifies candidate bifurcations. All other configurations are ignored. Representative neighborhood patterns and their corresponding encoded byte \(E(x,y)\) and \(\text{LUT}_{CN}\) values are illustrated in Table~\ref{tab:lut_examples}.

Boundary artifacts are eliminated using the Chebyshev distance transform \(D_M(x,y)\) of the segmentation mask \(M\). Candidates satisfying \(D_M(x,y) \le d_{\text{bg}}\) are pruned prior to ridge tracking.

\begin{table}[tb]
\centering
\caption{Representative \(3 \times 3\) neighborhood patterns, their encoded state \(E(x,y)\), \(CN\) values from \(\text{LUT}_{CN}\), and valid successor directions \(\mathcal{D}_{\text{next}}\) from \(\text{LUT}_{ND}\) for reachable prior directions \(d_{\text{prev}} \in \{0,\dots,7\}\) and start state (\(d_{\text{prev}}=8\)).}
\label{tab:lut_examples}
\small
\setlength{\tabcolsep}{3pt}
\begin{tabular}{cccl}
\toprule
Pattern & \(E(x,y)\) & \(\text{LUT}_{CN}\) & \(\text{LUT}_{ND}(d_{\text{prev}} \to [\mathcal{D}_{\text{next}}^1, \mathcal{D}_{\text{next}}^2, ...])\) \\

\midrule

\gridthree{0}{0}{0}{0}{1}{0}{0}{0}{0} & 0 & 0 & --- \\
\addlinespace[2pt]

\multirow{2}{*}{\gridthree{1}{0}{0}{0}{1}{0}{0}{0}{0}} & \multirow{2}{*}{1} & \multirow{2}{*}{1} & \(4 \to [\,]\) \\
 & & & \(8 \to [0]\) \\
\dots & \dots & \dots & \dots \\

\multirow{2}{*}{\gridthree{0}{0}{0}{0}{1}{0}{0}{1}{0}} & \multirow{2}{*}{32} & \multirow{2}{*}{1} & \(1 \to [\,]\) \\
 & & & \(8 \to [5]\) \\
\addlinespace[2pt]

\multirow{2}{*}{\gridthree{1}{0}{0}{0}{1}{0}{0}{1}{0}} & \multirow{2}{*}{33} & \multirow{2}{*}{2} & \(1 \to [0]\); \(4 \to [5]\) \\
 & & & \(8 \to [0, 5]\) \\
\dots & \dots & \dots & \dots \\

\multirow{2}{*}{\gridthree{1}{0}{1}{0}{1}{0}{0}{1}{0}} & \multirow{2}{*}{37} & \multirow{2}{*}{3} & \(1 \to [0, 2]\); \(4 \to [5, 2]\); \(6 \to [5, 0]\) \\
 & & & \(8 \to [0, 2, 5]\) \\
\dots & \dots & \dots & \dots \\

\bottomrule
\end{tabular}
\end{table}

\subsection{Skeleton tracking and direction estimation}
\label{subsec:skeleton_tracking}
Validated candidates undergo fast skeleton traversal to estimate their direction \(\theta \in [0, 2\pi)\) and initial quality score \(q_0\). Path traversal is modeled as a deterministic finite-state transition governed by a pre-computed state table \(\text{LUT}_{ND}[b][d_{\text{prev}}]\), where \(b = E(x,y)\) is the current neighborhood state and \(d_{\text{prev}} \in \{0, \dots, 7, 8\}\) encodes the entry direction (\(d_{\text{prev}} = 8\) denotes initial candidate departure). As shown in Table~\ref{tab:lut_examples}, \(\text{LUT}_{ND}\) directly returns an ordered list of valid successor directions \(\mathcal{D}_{\text{next}}\). This state transition mechanism intrinsically prevents 180-degree backtracking and orders candidate branches by minimal angular deflection relative to \(d_{\text{prev}}\), bypassing conditional branching during line tracking.

Tracking traverses the skeleton until an intersection/endpoint (\(CN \neq 2\)) is encountered or the cumulative path distance reaches \(d_{\text{max}}\). The minutia direction \(\theta\) and effective path length \(L\) are determined as follows:
\begin{itemize}
    \item \textit{Ridge endings:} \(L\) is the length of the single tracked path, and \(\theta\) is the orientation of the line segment connecting the candidate location to the final reached location.
    \item \textit{Bifurcations:} Tracking proceeds independently along all three departing branches. \(L\) is computed as the arithmetic mean of the three branch lengths, whereas \(\theta\) is defined as the bisecting direction of the two branches forming the acute enclosed angle.
\end{itemize}

Candidates with path length \(L < d_{\text{min}}\) are discarded as short noise artifacts. Validated minutiae receive an initial quality score proportional to their tracking extent, defined as \(q_0 = L / d_{\text{max}}\).

\subsection{Dual ridge-valley fusion}
\label{subsec:dual_extraction}
When dual-skeleton extraction is enabled, candidate detection and tracking are performed independently on both the ridge binary map \(B_R\) and the inverted valley binary map \(B_V\). To map valley minutiae \(\mathcal{M}_V\) into the ridge domain, their structural types are inverted (endings become bifurcations and vice versa), and their spatial coordinates are shifted by \(\delta_{\text{val}}\) pixels along (for endings) or opposite to (for bifurcations) their direction \(\theta\).

Bipartite matching then pairs candidates between ridge set \(\mathcal{M}_R\) and transformed valley set \(\mathcal{M}_V\) within a maximum distance tolerance \(d_{\text{pair}}\) and angular tolerance \(\theta_{\text{pair}}\). Unpaired minutiae from both domain sets are preserved to ensure high recall, but their quality scores are penalized by factor \(w_{\text{unpair}}\), yielding \(q = q_0 \cdot w_{\text{unpair}}\).

\subsection{Cluster-based quality decay}
\label{subsec:cluster_suppression}
High minutiae density typically indicates localized background noise, scars, or skeletonization artifacts. Rather than enforcing hard distance pruning—which risks removing legitimate closely-spaced minutiae—SBMEX applies a continuous quality attenuation based on local proximity.

For a set of extracted candidate minutiae \(\mathcal{M} = \{m_1, m_2, \dots\}\), the spatial proximity \(\Psi_{ij}\) between any pair of distinct minutiae \(m_i, m_j \in \mathcal{M}\) separated by Euclidean distance \(d_{ij} = \| (x_i, y_i) - (x_j, y_j) \|_2\) is modeled via a Gaussian kernel:
\begin{equation}
    \Psi_{ij} = \exp\left( -\frac{d_{ij}^2}{2\sigma_{\text{cl}}^2} \right)
\end{equation}
where \(\sigma_{\text{cl}}\) represents the cluster interaction radius. The cumulative density index for minutia \(m_i\) is given by \(D_i = \sum_{j \neq i} \Psi_{ij}\). The final adjusted quality score \(q_i^*\) is calculated via exponential decay: \(q_i^* = q_i \cdot \gamma_{\text{cl}}^{D_i}\), where \(\gamma_{\text{cl}} \in (0, 1]\) is the decay base factor. Highly dense clusters experience exponential quality attenuation, providing soft-scoring that downstream matchers can easily threshold.

\section{Experimental evaluation}
\label{sec:experiments}
This section evaluates the accuracy and computational efficiency of SBMEX in comparison with the three traditional open-source baselines described in Section~\ref{subsec:baselines}.

\subsection{Datasets and enhancement pipeline}
\label{subsec:datasets_pipeline}
\begin{figure}
	\centering
	\includegraphics[width=\columnwidth]{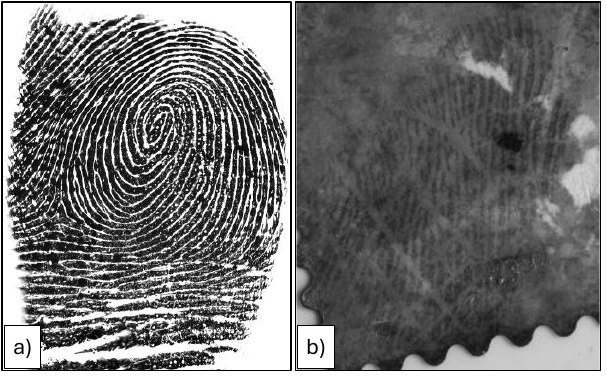}
	\caption{Representative samples from the NIST SD302 datasets: (a) high-quality optical exemplar impression from SD302g, and (b) challenging latent impression from SD302h showing partial ridge patterns and severe background noise.}
	\label{fig:datasets}
\end{figure}

Experiments are conducted on two distinct subsets of the NIST SD302 benchmark~\cite{Fiumara2019, Fiumara2026}:
\begin{itemize}
    \item \textit{Exemplars (NIST SD302g):} A collection of 2,380 high-quality exemplar fingerprints acquired at 1000 or 500\,DPI using optical sensors.
    \item \textit{Latents (NIST SD302h):} A challenging set of 3,815 latent impressions left naturally by subjects during simulated daily activities (e.g., handling paper currency or physical objects) and developed via standard forensic chemical and physical techniques. All images in this subset are categorized as ``value for comparison'' by latent examiners.
\end{itemize}
Both subsets feature ground-truth minutiae annotations meticulously validated by certified forensic examiners~\cite{Fiumara2026}. While the Latent subset is accompanied by manually annotated segmentation masks, those for Exemplar impressions are automatically generated via standard thresholding owing to their uniform white background. To standardize the evaluation across all experiments, all images are downsampled to a spatial resolution of 500\,DPI. Representative samples from both subsets are shown in Fig.~\ref{fig:datasets}.

For algorithms requiring an almost-binary enhanced input (namely SBMEX and CNBaseline), a traditional deterministic enhancement pipeline is deployed using the open-source \texttt{pyfing} ecosystem~\cite{Cappelli2024b}. Specifically, raw input images undergo four sequential processing stages: i) Ridge pattern segmentation using GMFS~\cite{Cappelli2023}; ii) Orientation field estimation via GBFOE~\cite{Cappelli2024}; iii) Spatial ridge frequency estimation via XSFFE~\cite{Cappelli2024a}; iv) Image enhancement using GBFEN~\cite{Cappelli2026}.

\subsection{Evaluation protocol}
\label{subsec:protocol}
To ensure fair and rigorous evaluation, the default parameters of SBMEX (summarized in Table~\ref{tab:sbmex_params}) were tuned offline on the FVC2000 DB2-A benchmark~\cite{Maio2002}. This dataset consists of fingerprint impressions acquired from untrained subjects without quality control, providing a representative baseline for parameter optimization. Crucially, FVC2000 DB2-A is completely disjoint from the NIST SD302 evaluation sets in terms of both subjects and acquisition sensor technologies. All experimental evaluations across both NIST SD302 subsets are conducted using these exact fixed parameters, without any dataset-specific fine-tuning.

The systematic evaluation protocol established in~\cite{Cappelli2026a} is adopted to assess intrinsic minutiae detection accuracy independently of downstream matching heuristics, which may mask detection errors (e.g., in position, direction, or type) via error-tolerant algorithms.
The protocol comprises two main evaluation steps:
\begin{enumerate}
    \item \textit{Minutiae pairing and filtering:} Extracted and ground-truth minutiae are paired by bipartite matching~\cite{Crouse2016}. A match is classified as a True Positive (TP) if the spatial distance and angular deviation are within specified thresholds \((\rho_t, \theta_t)\). Three increasingly stringent threshold levels are evaluated: ($\rho_t=$16$\,\text{px}, \theta_t=\pi/6\,\text{rad}$), ($12$\,px, \(\pi/8\)\,rad), and ($8$\,px, \(\pi/10\)\,rad). The evaluation region is restricted to the ground-truth segmentation bounding box, excluding a $14$\,px boundary margin to eliminate upstream segmentation biases (see~\cite{Cappelli2026a}).
    \item \textit{Aggregate Performance Metrics:} Precision-Recall (PR) curves are generated by sweeping the confidence threshold for extractors that produce quality scores, reporting the maximum achievable \(F_1\)-score. For CNBaseline, which yields unranked detections without confidence scores, performance is evaluated at its single operational point. Both \textit{type-agnostic} and \textit{type-aware} matching regimes are evaluated, where the latter requires matching minutia types for a TP (see~\cite{Cappelli2026a}).
\end{enumerate}

\subsection{Results and discussion}
\label{subsec:results}
Table~\ref{tab:results_sd302_combined} presents the quantitative comparison (\(F_1\)-scores across the three threshold levels) on the NIST SD302 benchmark datasets under both type-agnostic and type-aware matching regimes. To isolate feature extraction performance from pre-processing effects, all extractors capable of processing grayscale images—encompassing both traditional baselines (MINDTCT, FingerJet) and state-of-the-art methods (VeriFinger, LEADER)—are evaluated on both original (\textit{Raw}) inputs and pre-processed (\textit{Enh}) inputs enhanced via the \texttt{pyfing} pipeline (Section~\ref{subsec:datasets_pipeline})\footnote{Since the \texttt{pyfing} pipeline yields enhanced binary images with white ridges on a black background, enhanced inputs passed to these extractors were inverted to match their expected input conventions.}. Although passing enhanced images to off-the-shelf tools may induce redundant internal filtering steps not originally intended by their design, it provides the most rigorous way to evaluate these tools ``as-is'' under identical pre-processing conditions. Commercial (VeriFinger) and deep learning (LEADER) methods are included on both inputs for completeness, serving as external benchmarks for context, and are excluded from the traditional baseline ranking. All reported \(F_1\)-score differences are statistically robust; \(95\%\) confidence intervals estimated via 1,000 non-parametric image-level bootstrap resamples were omitted from Table~\ref{tab:results_sd302_combined} for space constraints, as they all fall within \(\le 0.01\).

On Exemplar impressions, SBMEX achieves top-ranking performance among traditional open-source tools. In the type-agnostic regime, SBMEX consistently outperforms all traditional competitors across all threshold levels, achieving an \(F_1\)-score of $0.68$ at ($16$\,px, \(\pi/6\)\,rad). Under the type-aware regime, SBMEX remains superior or competitive, matched only by MINDTCT with enhanced input. Notably, at the most stringent thresholds ($8$\,px, \(\pi/10\)\,rad), SBMEX ($0.63$ type-agnostic) substantially narrows the performance gap with commercial COTS systems such as VeriFinger ($0.64$ on Raw input). On the challenging Latent subset, performance drops markedly across all traditional extractors due to severe noise. Nevertheless, SBMEX and its ridge-only variant SBMEX-R demonstrate superior robustness, outperforming all traditional open-source baselines by a significant margin in both type-agnostic and type-aware regimes. Across both datasets, SBMEX-R exhibits only a slight performance decay compared to the full dual-skeleton SBMEX, establishing SBMEX-R as a highly effective option when processing resources are constrained.

\begin{figure*}
\centering
\begin{tikzpicture}
    \begin{groupplot}[
        group style={
            group size=4 by 2,
            horizontal sep=0.15cm,
            vertical sep=0.25cm,
            x descriptions at=edge bottom,
            y descriptions at=edge left,
        },
        plotstyle,
        width=5.5cm,
        height=5cm,
        xmin=0, xmax=1, ymin=0, ymax=1,
        grid=both,
        xtick={0,0.1,...,1},
        ytick={0,0.1,...,1},
        cycle list name=graph_precision_recall,
    ]
    \input{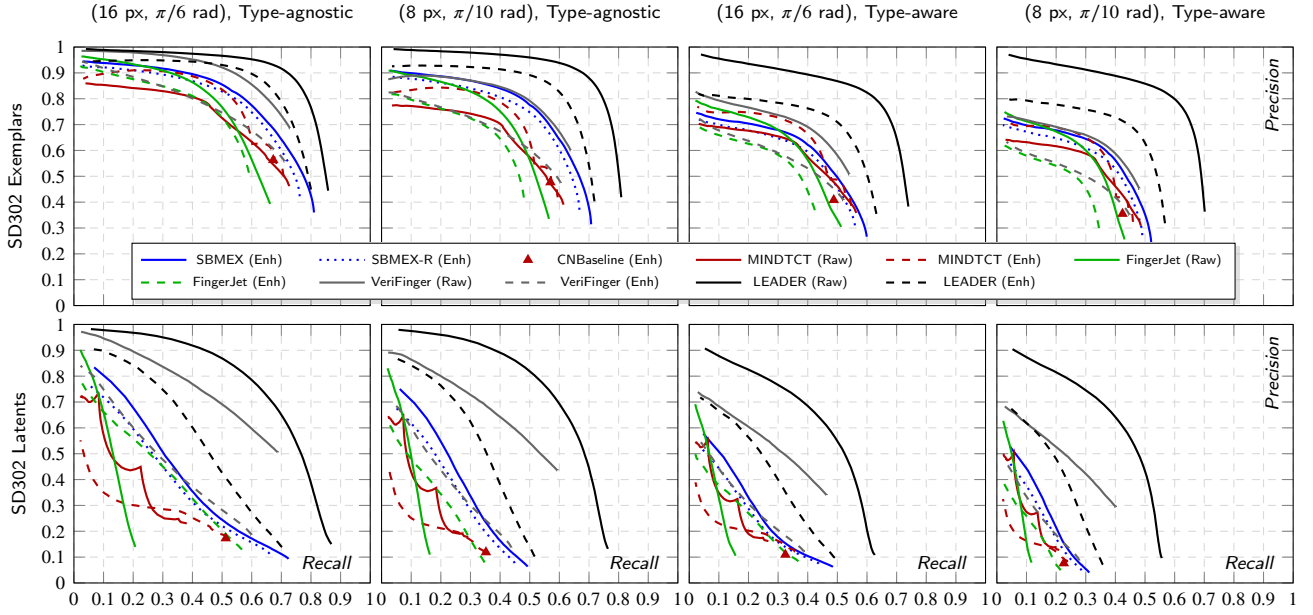}
    \end{groupplot}
    \coordinate (top_row_center) at ($(group c1r1.south)!.5!(group c4r1.south)$);
    \node[above, yshift=0.07cm,
        fill=white,
        drop shadow={
            shadow xshift=0.4ex,
            shadow yshift=-0.4ex,
            fill=black!30,
            opacity=0.4
        }, inner sep=0pt] at (top_row_center) {\pgfplotslegendfromname{single_legend}};
\end{tikzpicture}
\caption{PR curves on the two datasets across two threshold levels under type-agnostic and type-aware regimes.}
\label{fig:pr_curves}
\end{figure*}

Figure~\ref{fig:pr_curves} depicts the Precision-Recall (PR) curves across two representative threshold levels. The PR trajectories confirm the findings from Table~\ref{tab:results_sd302_combined}: the curve for SBMEX (and SBMEX-R running closely parallel just below it) consistently dominates those of traditional baselines across almost the entire operational range. Minor exceptions occur on Exemplars in the type-aware regime at the ($16$\,px, \(\pi/6\)\,rad) threshold level, where MINDTCT (Enh) and FingerJet (Raw) gain a localized advantage at low recall operating points.

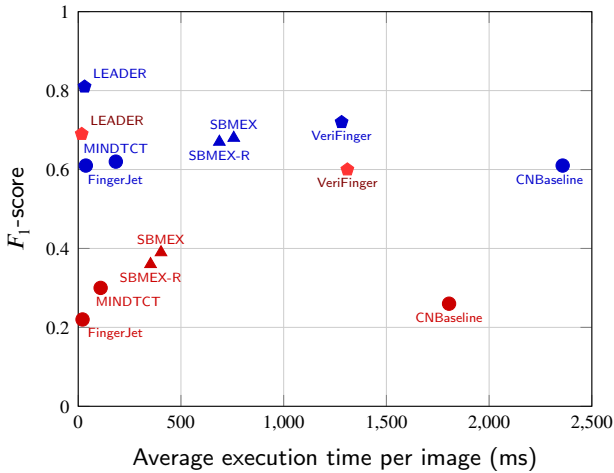
\begin{figure}
\centering
\begin{tikzpicture}[scale=1.0]
\begin{axis}[
    width=\linewidth,
    height=6.8cm,
    %xmode=log,
    log basis x={10},
    xmin=0, xmax=2500,
    ymin=0, ymax=1.0,
    xlabel={Average execution time per image (ms)},
    ylabel={\(F_1\)-score},
    grid=both,
    grid style={line width=.1pt, draw=gray!20},
    major grid style={line width=.3pt, draw=gray!40},
    legend pos=south east,
    legend style={font=\tiny, fill=white, fill opacity=0.85, draw opacity=1, cells={anchor=west}},
    tick label style={font=\scriptsize},
    label style={font=\small},
    xminorticks=true
]

\addplot[only marks, mark=*, mark size=2.5pt, color=blue!80!black] coordinates {(37, 0.61)};
\node[anchor=north west, font=\tiny, color=blue!80!black] at (axis cs:5, 0.61) {FingerJet};

\addplot[only marks, mark=*, mark size=2.5pt, color=blue!80!black] coordinates {(183, 0.62)};
\node[anchor=south, font=\tiny, color=blue!80!black] at (axis cs:183, 0.62) {MINDTCT};

\addplot[only marks, mark=pentagon*, mark size=2.5pt, color=blue!80!black] coordinates {(1282, 0.72)};
\node[anchor=north, font=\tiny, color=blue!80!black] at (axis cs:1282, 0.72) {VeriFinger};

\addplot[only marks, mark=pentagon*, mark size=2.5pt, color=blue!80!black] coordinates {(31, 0.81)};
\node[anchor=south west, font=\tiny, color=blue!80!black] at (axis cs:31, 0.81) {LEADER};

\addplot[only marks, mark=*, mark size=2.5pt, color=blue!80!black] coordinates {(2358, 0.61)};
\node[anchor=north east, font=\tiny, color=blue!80!black] at (axis cs:2500, 0.61) {CNBaseline};

\addplot[only marks, mark=triangle*, mark size=2.5pt, color=blue!80!black] coordinates {(757, 0.68)};
\node[anchor=south, font=\tiny, color=blue!80!black] at (axis cs:757, 0.68) {SBMEX};

\addplot[only marks, mark=triangle*, mark size=2.5pt, color=blue!80!black] coordinates {(687, 0.67)};
\node[anchor=north, font=\tiny, color=blue!80!black] at (axis cs:687, 0.67) {SBMEX-R};

\addplot[only marks, mark=*, mark size=2.5pt, color=red!80!black] coordinates {(21, 0.22)};
\node[anchor=north west, font=\tiny, color=red!80!black] at (axis cs:5, 0.22) {FingerJet};

\addplot[only marks, mark=*, mark size=2.5pt, color=red!80!black] coordinates {(109, 0.30)};
\node[anchor=north west, font=\tiny, color=red!80!black] at (axis cs:44, 0.30) {MINDTCT};

\addplot[only marks, mark=*, mark size=2.5pt, color=red!80!black] coordinates {(1805, 0.26)};
\node[anchor=north, font=\tiny, color=red!80!black] at (axis cs:1805, 0.26) {CNBaseline};

\addplot[only marks, mark=triangle*, mark size=2.5pt, color=red!80!black] coordinates {(352, 0.36)};
\node[anchor=north, font=\tiny, color=red!80!black] at (axis cs:352, 0.36) {SBMEX-R};

\addplot[only marks, mark=triangle*, mark size=2.5pt, color=red!80!black] coordinates {(403, 0.39)};
\node[anchor=south, font=\tiny, color=red!80!black] at (axis cs:403, 0.39) {SBMEX};

\addplot[only marks, mark=pentagon*, mark size=2.5pt, color=red!80] coordinates {(1310, 0.60)};
\node[anchor=north, font=\tiny, color=red!50!black] at (axis cs:1310, 0.60) {VeriFinger};

\addplot[only marks, mark=pentagon*, mark size=2.5pt, color=red!80] coordinates {(17, 0.69)};
\node[anchor=south west, font=\tiny, color=red!50!black] at (axis cs:17, 0.69) {LEADER};

\end{axis}
\end{tikzpicture}
\caption{Accuracy vs. Latency trade-off, at ($16$\,px, \(\pi/6\)\,rad), type-agnostic, on Exemplars (blue) and Latents (red).}
\label{fig:f1_vs_time}
\end{figure}

Figure~\ref{fig:f1_vs_time} illustrates the trade-off between extraction accuracy ($F_1$-score) and execution latency measured at the ($16$\,px, $\pi/6$\,rad) threshold level under the type-agnostic regime. Benchmark timings were recorded on an Intel\textregistered\ Xeon\textregistered\ Silver 4112 CPU @ 2.60\,GHz, with the exception of LEADER (executed on GPU). For SBMEX, SBMEX-R, and CN Baseline, reported latencies explicitly encompass the full end-to-end pipeline including \texttt{pyfing} pre-processing, whereas other baselines are measured on original raw inputs. SBMEX achieves an average total execution time of $757$\,ms on Exemplars and $403$\,ms on Latents. Although legacy C/C++ compiled packages such as MINDTCT ($183$\,ms / $109$\,ms) and FingerJet ($37$\,ms / $21$\,ms) exhibit lower processing latency on raw inputs, this speed comes at the cost of substantially lower extraction accuracy, particularly on noisy Latents where SBMEX provides a marked $F_1$-score improvement ($0.39$ vs. $0.30$ and $0.22$). Conversely, SBMEX is considerably faster than the commercial COTS engine VeriFinger ($1282$\,ms / $1310$\,ms) and drastically outperforms CNBaseline ($2358$\,ms / $1805$\,ms), the only other Python-native baseline. Notably, when isolating pure minutiae detection by subtracting the \texttt{pyfing} pre-processing latency ($645$\,ms for Exemplars and $322$\,ms for Latents), the core detection time of SBMEX drops to just $112$\,ms ($81$\,ms on Latents), while the lightweight SBMEX-R reaches $42$\,ms ($30$\,ms on Latents). In comparison, CNBaseline requires $1713$\,ms and $1483$\,ms for core detection alone, underscoring the superior efficiency-accuracy balance achieved by the LUT-based architecture of SBMEX.

\newcommand{\thrl}[2]{\ensuremath{\substack{#1\,\text{px}\\\pi/#2\,\text{rad}}}}

\begin{table*}[!t]
\centering
\caption{Comparative performance (\(F_1\)-score) on the two datasets across the three threshold levels under type-agnostic and type-aware regimes. \textbf{Bold} indicates the best performance among competing methods, while \underline{underline} denotes the second best.}
\label{tab:results_sd302_combined}
\begin{threeparttable}
\scriptsize
\setlength{\tabcolsep}{2pt}
\begin{tabular*}{\textwidth}{@{\extracolsep{\fill}} lc ccc c ccc c ccc c ccc @{}}
\toprule
& & \multicolumn{7}{c}{\textbf{NIST SD302 Exemplar}} & & \multicolumn{7}{c}{\textbf{NIST SD302 Latent}} \\
\cmidrule(lr){3-9} \cmidrule(lr){11-17}
& & \multicolumn{3}{c}{Type-agnostic} & & \multicolumn{3}{c}{Type-aware} & & \multicolumn{3}{c}{Type-agnostic} & & \multicolumn{3}{c}{Type-aware} \\
\cmidrule(lr){3-5} \cmidrule(lr){7-9} \cmidrule(lr){11-13} \cmidrule(lr){15-17}
\textbf{Method} & \textbf{Input} & \thrl{16}{6} & \thrl{12}{8} & \thrl{8}{10} & & 
                   \thrl{16}{6} & \thrl{12}{8} & \thrl{8}{10} & & 
                   \thrl{16}{6} & \thrl{12}{8} & \thrl{8}{10} & & 
                   \thrl{16}{6} & \thrl{12}{8} & \thrl{8}{10} \\
\midrule

% --- Copiare qui sotto

SBMEX & Enh & \textbf{0.68} & \textbf{0.66} & \textbf{0.63} & & \underline{0.50} & \textbf{0.49} & \textbf{0.47} & & \textbf{0.39} & \textbf{0.36} & \textbf{0.32} & & \textbf{0.25} & \textbf{0.24} & \textbf{0.21} \\
SBMEX-R & Enh & \underline{0.67} & \underline{0.65} & \underline{0.61} & & 0.49 & 0.48 & \underline{0.46} & & 0.36 & \underline{0.34} & \underline{0.30} & & \underline{0.24} & \underline{0.22} & \underline{0.20} \\
CNBaseline & Enh & 0.61 & 0.58 & 0.52 & & 0.44 & 0.42 & 0.39 & & 0.26 & 0.22 & 0.18 & & 0.16 & 0.14 & 0.11 \\
MINDTCT & Raw & 0.62 & 0.59 & 0.53 & & 0.48 & 0.46 & 0.43 & & 0.30 & 0.28 & 0.25 & & 0.22 & 0.20 & 0.18 \\
MINDTCT & Enh & 0.65 & 0.62 & 0.57 & & \textbf{0.51} & \textbf{0.49} & 0.45 & & 0.31 & 0.28 & 0.22 & & 0.22 & 0.19 & 0.15 \\
FingerJet & Raw & 0.61 & 0.59 & 0.54 & & 0.47 & 0.45 & 0.42 & & 0.22 & 0.20 & 0.18 & & 0.16 & 0.15 & 0.13 \\
FingerJet & Enh & 0.60 & 0.56 & 0.50 & & 0.43 & 0.40 & 0.36 & & \underline{0.37} & 0.33 & 0.26 & & 0.23 & 0.21 & 0.16 \\
\hline
\multicolumn{17}{l}{\textit{Commercial and Deep Learning SOTA Methods}} \\
VeriFinger & Raw & 0.72 & 0.69 & 0.64 & & 0.54 & 0.52 & 0.48 & & 0.60 & 0.57 & 0.52 & & 0.41 & 0.39 & 0.36 \\
VeriFinger & Enh & 0.64 & 0.61 & 0.55 & & 0.47 & 0.45 & 0.41 & & 0.39 & 0.35 & 0.30 & & 0.25 & 0.23 & 0.19 \\
LEADER & Raw & 0.81 & 0.80 & 0.78 & & 0.71 & 0.70 & 0.69 & & 0.69 & 0.67 & 0.64 & & 0.53 & 0.52 & 0.50 \\
LEADER & Enh & 0.74 & 0.73 & 0.70 & & 0.59 & 0.58 & 0.56 & & 0.48 & 0.46 & 0.42 & & 0.34 & 0.33 & 0.30 \\

% --- Copiare qui sopra
\bottomrule
\end{tabular*}
\end{threeparttable}
\end{table*}

%-----------------------------------------------------------------------------

%-------------------------------------------------------------------------------

\section{Conclusion}
\label{sec:conclusion}
This paper presented SBMEX, a deterministic, skeleton-based minutiae detection method designed to provide a modern, transparent baseline for standard fingerprint pipelines. By combining LUT-accelerated skeleton processing, dual-skeleton fusion, and continuous quality decay, SBMEX eliminates runtime neighborhood scans while maintaining high accuracy and algorithmic explainability. Experimental evaluations on the NIST SD302 benchmark demonstrate that SBMEX achieves minutiae detection performance fully comparable to legacy traditional baselines, while delivering a drastic reduction in core processing time relative to a classical CN Python implementation.

Rather than attempting to supersede end-to-end deep learning architectures or commercial matchers in unconstrained scenarios, SBMEX addresses the ongoing need for handcrafted pipelines that offer full auditability, standard compliance, and operational independence from GPUs or extensive training data. To bridge the gap between legacy C/C++ tools and modern scientific workflows, SBMEX is fully integrated into the open-source \texttt{pyfing} ecosystem~\cite{Cappelli2024b}. Future research will focus on a comprehensive empirical study comparing fully traditional, deep learning-based, and hybrid pipelines across diverse operational conditions.

\small
\setlength{\bibsep}{4pt}
\bibliographystyle{unsrtnat}
\bibliography{fingerprint}

\end{document}